\documentclass[a4paper]{article}
\usepackage{graphicx}
\usepackage{amssymb}
\usepackage{amsmath}
\usepackage{caption}

\usepackage[pagebackref=true,breaklinks=true]{hyperref}

\date{}

\begin{document}

\title{Comparison of Methods Generalizing Max- and Average-Pooling}
\author{Florentin~Bieder, Robin~Sandk\"uhler, Philippe~C.~Cattin\\~\\
{\small Department of Biomedical Engineering, University of Basel, Switzerland}\\
{\small \href{mailto:florentin.bieder@unibas.ch}{\tt florentin.bieder@unibas.ch}}
}

\maketitle

\begin{abstract}
Max- and average-pooling are the most popular pooling methods for 
downsampling in convolutional neural networks. In this paper, 
we compare different pooling methods that generalize both max- 
and average-pooling. Furthermore, we propose another method 
based on a smooth approximation of the maximum function and 
put it into context with related methods. 
For the comparison, we use a VGG16 
image classification network and train it on a large dataset 
of natural high-resolution images (Google Open Images v5). 

The results show that none of the more sophisticated methods 
perform significantly better in this classification task 
than standard max- or average-pooling.
\end{abstract}

\section{Introduction}
\label{introduction}
In the design of convolutional neural networks (\emph{CNN}s), 
pooling blocks are important elements that are used for 
downsampling feature maps.

Firstly, downsampling reduces the spatial resolution of the 
feature maps, which is desirable when we want to extract semantic 
information. Classification networks usually apply this idea repeatedly.

Secondly, the reduction of the resolution can be advantageous when 
considering the cost in resources: 
For optimizing of the parameters of a network, 
gradient-based optimizers are used. The most common way 
to compute the gradient is the backpropagation algorithm. 
It requires the intermediate feature maps from evaluating 
the network to be retained for the subsequent computation of the gradient. 
The memory used for saving them is one of the limiting factors of the current hardware. By reducing the resolution, we can reduce the amount of memory needed.

It is worth noting that while pooling methods are 
frequently used for downsampling, most of them can also be used 
in ways that preserve the size of the feature maps. In this case, 
they can for instance also act as a convolutional block 
(which we do in fact also consider as a pooling method), as a nonlinearity 
(e.g. max-pooling) or as a blurring block (e.g. average-pooling). 
In this paper, we will focus on the application of pooling as a 
downsampling method.

The max-pooling technique was already used by Weng et al. in the \emph{Cresceptron} 
\cite{weng}, and is still one of the most popular pooling 
methods today. Another popular method is average-pooling: 
When using average-pooling for halving the feature map size, 
we can consider it a special case of bilinear downsampling.
Average-pooling was used by LeCun et al. in the seminal \emph{LeNet} \cite{lecun}, inspired by the Fukushima's \emph{Neocognitron}~\cite{fukushima}.

While max-pooling makes a choice (choosing the maximal value) 
and preserves the most prominent features, 
average-pooling has a smoothing effect. Both of them have been 
claimed to reduce the sensitivity to translation within the input. But it was shown that the number of pooling blocks only plays 
a secondary role when it comes to making the networks insensitive 
to the translation of features in the input, as shown by Kauderer-Abrams \cite{abrams}.

In the context of CNNs, we also 
find strided convolutions being used for downsampling. 
As we will discuss later, these can be considered as usual 
convolutions followed by a nearest-neighbor-downsampling. Just as in 
max-pooling, only the value of a certain position is propagated, 
and the rest is ignored.

This is also reflected in the gradient: Given a neighborhood window of pixels ${x \in \mathbb R^n}$, both nearest-neighbor-downsampling and max-pooling ``choose'' a certain pixel $x_j$ to be propagated. In the case of nearest-neighbor-downsampling $j$ is chosen based on the location, and in max-pooling $j$ is chosen by the relative values (i.e. ${j= \arg\max_i x_i}$, assuming the maximum is unique). Let us define the function $f$ as the action that nearest-neighbor-downsampling or max-pooling performs on that given neighborhood $x$. The gradient is then given by  

\begin{equation}
(\nabla f)_i  = \begin{cases} 
1 & \text{ if } i = j
\\ 0 &  \text{ otherwise}
\end{cases}.
\end{equation}

Therefore we get non-zero gradient information for only \emph{one} of the arguments.\footnote{Assuming the maximum is unique. The gradient is not defined if the maximum is attained by more than one argument. Since this is just a zero set of the input space, many implementations just set the gradient to $0$ or $1$ in this case.}
In contrast, the gradient of an average function ${f(x) = \frac 1n \sum_i x_i}$, which is applied to each neighborhood in average-pooling, is

\begin{equation}
 (\nabla f )_j = \frac 1n \qquad \forall j, 
\end{equation}
and therefore propagates non-zero gradient information into \emph{all} arguments. 
This statement can also be extended to any weighted average with non-zero weights.
It is, however, unclear what implications these differences between these methods have.

Many different ways to generalize average- and max-pooling have 
been suggested in an attempt to harness the advantages of both methods, 
and to improve the expressibility of the network \cite{lee, deliege, gulcehre, pinheiro}.

They all have in common that they have a small number of additional 
parameters that allow for a smooth transition between average- 
and max-pooling. It should be noted that some of the methods 
can attain the exact max- and average-pooling behaviour 
only in the limit, but for practical purposes this is not a limitation, 
as for sufficiently extreme values of the parameters, the 
differences are in the order of machine precision.

In this paper, we investigate how the different methods 
that generalize both ideas of max- and average-pooling 
perform in a classification task of natural, high-resolution images. 
Furthermore, we propose another pooling method that also generalizes 
average- and max-pooling and compare it to the others.

\section{Background}
\subsection{Pooling}
In most CNN architectures, pooling operations are applied 
to reduce the spatial resolution of the feature maps. 
We will consider the case of 2D images, but all methods can be used for any number of dimensions, for instance 1D, when using sound or other temporal data, or 3D when processing MRT or CT scans.

\subsection{Notation}

For a given pooling operation we denote the input as 
$X \in \mathbb R^{C\times H \times W}$ 
and the output as ${Y \in \mathbb R^{C' \times H' \times W'}}$. Here $H$ and $W$ denote the height and width of the feature maps, and $C$ denotes the number of channels.
For simplicity, we set $C = C'$ and omit the corresponding index $c$, 
unless noted otherwise.
The pooling methods we consider are all \emph{sliding window methods}. In a sliding window method with a window size of $k_1 \times k_2$ 
and stride lengths of $s_1 \times s_2$, for each $i,j$ we consider the window 
$\mathcal W_{ij}\colon  \mathbb R^{H \times W} \to \mathbb R^{k_1 \times k_2} $ with

\begin{equation}
\mathcal W_{ij} (X) =  
\begin{pmatrix}
X_{s_1(i-1)+1,s_2(j-1)+1} & \ldots & X_{s_1(i-1)+k_1,s_2(j-1)+1}\\
\vdots &  & \vdots \\
X_{s_1(i-1)+1,s_2(j-1)+k} & \ldots & X_{s_1(i-1)+k_1,s_1(j-1)+k_2}
\end{pmatrix}.
\end{equation}

Any sliding window method that maps $X$ to $Y$ can be defined for each entry $Y_{ij}$ as

\begin{equation}
Y_{ij} 
= 
f(\mathcal W_{ij}(X)) 
\end{equation}
for some appropriate function $f\colon \mathbb R^{k_1 \times k_2} \to \mathbb R$.

\begin{figure}[htbp]
\begin{center}
\centerline{\includegraphics[scale=1]{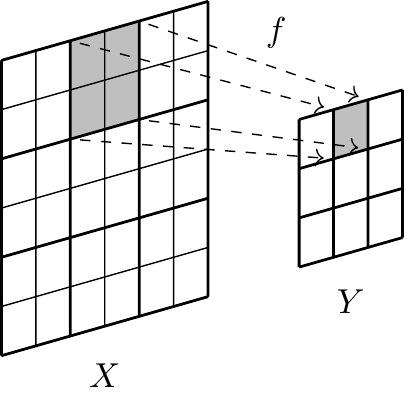}}
\caption{Visualization of a 2D pooling block with a window size of $2 \times 2$ and a stride of $2\times 2$ with $H=W=6$.}
\label{fig:slidingwindow}
\end{center}
\end{figure}

The output $Y$ will have the size

\begin{equation}
(H', W') = \left( 
\left[ \frac{H-k_1}{s_1}\right] + 1,
\left[ \frac{W-k_2}{s_2}\right] + 1,
\right),
\end{equation}
where the Gauss-brackets $[ t ] \in \mathbb Z$ denote the greatest integer that is not greater than $t \in \mathbb R$. Figure \ref{fig:slidingwindow} shows a visualization of a generic pooling operation.

Each sliding window is processed the same way and is independent of the others. From now on, we will therefore just write $x = \mathcal W_{ij}(X)$ for the sliding window. For simplicity, we rewrite the indices $x = (x_1, \ldots, x_n)$ where $n$ is the number of entries in the window. 
The following section will introduce several different pooling functions $f$.

\subsubsection{Pooling Methods}
Let us consider square input- and output images $X$ and $Y$ with $H=W$ and $H' = W'$. In a typical setting, windows of size $k \times k$ with a stride of $s=k$ (in both directions) are used to scale the image by a factor of $1/k$. Most of the time a factor of $1/2$ is used, which means a $2\times 2$ window is used and $H = 2H'$.

\paragraph{Max-Pooling and Average-Pooling} With this notation, max- pooling (\texttt{MP}) can be written as

\begin{equation} 
f_{\operatorname{MP}}(x) 
= 
\max_i x_i. 
\end{equation}

For the very similar average-pooling (\texttt{AP}), we have

\begin{equation} 
f_{\operatorname{AP}}(x) 
= 
\frac{1}{n} \sum_{i=1}^n x_i.
\end{equation}

\paragraph{Convolutions}
Convolutions with weights $w_i$ can be implemented with 
\begin{equation} 
f_{\operatorname{CONV}}^w(x) 
= 
\sum_{i=1}^{n} w_i x_i. 
\end{equation}

It is worth noting that for downsampling we can use \emph{strided convolutions}. 
And we can reformulate this by considering $X$ as the output of a standard convolution 
(i.e. with stride $1$), but then applying

\paragraph{Nearest-Neighbor-Downsampling}
We can formulate the Nearest-Neighbor-Downsampling (\texttt{NN}) with 

\begin{equation} 
f_{\operatorname{NN}}(x) = x_1 
\end{equation}

for the case of $k=2$. Actually, one would usually pick the central entry, but since there is no central entry in a $2\times 2$ window, we can choose any one of them, in this case, $x_1$.

\paragraph{Gated-Pooling} Introduced by Lee et al. \cite{lee} the \emph{gated-pooling} (\texttt{GP}) is defined as

\begin{equation}
f_{\operatorname{GP}}^w(x) = g\frac{1}{n}\sum_{i=1}^n x_i + (1-g)\max_i x_i,
\end{equation}
where $g = \sigma(w_1 x_1 + \ldots + w_n x_n)$ with trainable weights $w_i$ and the sigmoid function $\sigma(t) = (1+e^{-t})^{-1}$. In the case of more than one channel, the same weights are reused for all channels. Their experiments showed \texttt{GP} to be superior to \emph{mixed-pooling} \cite{lee} where $g$ is just a trainable parameter.

\paragraph{Ordinal-Pooling}
Introduced by Deli{\`e}ge et al. \cite{deliege}, \emph{ordinal-pooling} (\texttt{OP}) is a method that allows interpolating 
between the minimum and maximum. As the name suggests, 
the values within the window are sorted, and then reduced 
to one value using a trainable convex combination,

\begin{equation}\label{eq:fop}
f_{\operatorname{OP}}^w(x) 
= \sum_{i=1}^n w_i x_{\pi_{x_1,\ldots, x_n}(i)}
\end{equation}

where $\pi_{x_1, \ldots, x_n}\colon \{1,\ldots, n\} \to \{1, \ldots, n\}$ 
is a permutation that sorts the values $x_i$ in ascending order. The $w_i$ are trainable weights. Additionally, after each (gradient-based) optimization step, negative values are set to zero, and then they are normalized to sum up to $1$ with following update:

\begin{equation}
\tilde w_i  := \frac{\operatorname{ReLU}(w_i)}{\sum_j \operatorname{ReLU}(w_j)} \qquad \forall i = 1, \ldots, n
\end{equation}

Here $\tilde w$ denotes the weights after the update. In the case of more than one channel, the same weights are reused for all channels.

\paragraph{Learned-Norm-Pooling}
If the input $x$ consists of just positive values, 
then \texttt{AP} as well as \texttt{MP} can be considered as applying the $1$- or $\infty$-norm (up to some constant factor), 
both of which are special cases of the well known $\ell^p$-norms.

Gulcehre et al. \cite{gulcehre} proposed \emph{learned norm pooling} (\texttt{LNP}), where they use the $\ell^p$-norm with a trainable parameter $p > 1$, which can be formulated as 

\begin{equation}
f_{\operatorname{LNP}}^p(x) = \sqrt[p]{ \frac{1}{n} \sum_{i=1}^n \vert x_i \vert^p}.
\end{equation}

The authors proposed to parametrize $p$ with a new parameter $\tilde p \in \mathbb R$ as

\begin{equation}\label{eq:flnptilde}
p = 1 + \log( 1 + \exp(\tilde p))
\end{equation}
to ensure that $p$ remains in the range $(1,\infty)$. Note that contrary to the usual definition of the $\ell^p$-norm, \texttt{LNP} uses the \emph{average} instead of a \emph{sum}. 
Furthermore, the absolute value can be removed if the \texttt{LNP}-block directly follows a $\operatorname{ReLU}$ activation. 
This means that \texttt{LNP} only generalizes \texttt{MP} and \texttt{AP} if the arguments are nonnegative. If there are negative arguments present, $f_{\operatorname{LNP}}^p$ does \emph{not} coincide with $f_{\operatorname{MP}}$ or $f_{\operatorname{AP}}$ for any value of $p$. 
Finally, we would like to point out that in general, $f_{\operatorname{LNP}}^p$ has a singularity wherever $x_i = 0$. In such a case we just define an adjacent value of the derivative just as it is done for differentiating the $\operatorname{ReLU}$ activation function.

\paragraph{LSE-Pooling}
The \emph{log-sum-exp-pooling} (\texttt{LSE}) was introduced by Pinheiro and Ronan \cite{pinheiro}. It uses another well known smooth approximation of the maximum. This approximation is one of the family of \emph{quasi-arithmetic-} or \emph{Kolmogorov-}\emph{means} \cite{kolmogorov}:

\begin{equation}
f_{\operatorname{LSE}}^r(x) = \frac{1}{r} \log\left(\frac{1}{n} \sum_{i=1}^n \exp(r x_i) \right)
\end{equation}

where $r > 0$ is a parameter. For $r\to \infty$ this expression converges to the maximum, and for $r \to 0$ it converges to the average. 
Pinheiro and Ronan \cite{pinheiro} manually selected $r$, but did not specify the value they had selected. 
Furthermore, there are numerical issues in the computation of the gradient with respect to $x_i$ that were not addressed in the paper.
Therefore, we did not run any experiments with this method, 
but we still want to mention it due to its connections to other methods we used. In fact, it is closely related to the \emph{softmax}, as that is its gradient, given by

\begin{equation} \label{eq:flnp}
\frac{\partial  f_{\operatorname{LSE}}^r( x)}{\partial x_i}  = \frac{e^{rx_i}}{\sum_j e^{rx_j}} = \operatorname{softmax}(rx_1,\ldots, rx_n).
\end{equation}

\section{Method}

Besides comparing of existing pooling operations, we will now introduce a novel pooling method, which is based on a smooth approximation of the maximum. We will also point out connections to other functions such as the ``softmax'' and the ``log-sum-exp''-approximation.

\paragraph{Smooth Approximation of the Maximum}

The ``softmax'' (or ``softargmax'') is defined as

\begin{equation} 
\operatorname{softmax}(x) 
= 
\left( \frac{ e^{x_1}}{\sum_{j=1}^n e^{x_j}}, \ldots, \frac{ e^{x_n}}{\sum_{j=1}^n e^{x_j}} \right) .
\end{equation}

It provides a categorical probability distribution, with the greatest probability value for the greatest value of $x_i$. In that sense, it is a ``soft'' version of the one-hot-encoding of the argument of the maximum. It can be generalized by introducing a parameter $\tau$ as follows:

\begin{equation} 
\operatorname{softmax}_\tau(x) 
= 
\left( \frac{ e^{\tau x_1}}{\sum_{j=1}^n e^{\tau x_j}}, \ldots, \frac{ e^{\tau x_n}}{\sum_{j=1}^n e^{\tau x_j}} \right) .
\end{equation}

In physics, it is known as the \emph{Boltzmann-} or \emph{Gibbs-} distribution, 
where $\tau = -1/(kT)$ with $k$ the Boltzmann constant and $T$ the temperature. That is the reason why this parameter (sometimes also with a slightly different parametrization of $\pm 1/\tau$) is sometimes referred to as ``temperature''-parameter \cite{hinton}.

We would like to point out that the softmax takes the role of a smooth approximation of the \emph{arguments} of the maxima ($\arg\max$) and not of the maxima ($\max$). 
It is, however, closely related to the \emph{smooth maximum}, which is defined as

\begin{equation} 
\operatorname{smoothmax}(x) 
= 
\sum_{i=1}^n x_i \frac{  e^{x_i}}{\sum_{j=1}^n e^{x_j}}. 
\end{equation}

This is a convex combination of the arguments $x_i$ since the coefficients are all nonnegative and sum to one. 

In the following section, we will propose a method that - to the best of our knowledge - 
has not been studied in the context of downsampling operations for CNNs so far. A method for globally pooling features over a variable 1D- time domain, which is based on the same approximation of the smooth maximum, has been proposed by McFee et al. \cite{fee} for sound event detection. Furthermore the ``swish''-activation function 

\begin{equation}
\operatorname{swish}(x) = \frac{x}{1 + e^{-x}}
\end{equation} 
proposed by Ramachandran et al. \cite{ramachandran} can be derived from a $\operatorname{ReLU}(x) = \max(0,x)$ activation function, if we replace the maximum with the smooth maximum.

\paragraph{Smooth-Maximum-Pooling}

This smooth maximum function (just like softmax) can be generalized by introducing a 
parameter $\tau \in \mathbb R$: We propose to use 

\begin{equation} \label{eq:fsmp}
f_{\operatorname{SMP}}^\tau(x) 
:= 
\sum_{i=1}^n x_i \frac{e^{\tau x_i}}{\sum_{j=1}^n e^{\tau x_j}},
\end{equation}
as a pooling function, which generalizes this concept, and we call it \emph{smooth-maximum-pooling} (\texttt{SMP}). 
It is again a convex combination of the 
arguments and includes 
\begin{itemize}
\item[--] \texttt{AP} for $\tau = 0$, 
\item[--] \texttt{MP} for $\tau \to \infty$ and 
\item[--] min-pooling for $\tau \to -\infty$. 
\end{itemize}

The theoretical implementation details are documented in the appendix.

For each channel in the input, we use a separate value of $\tau$.
We propose three variations:
\begin{itemize}
\item[--] A fixed value $\tau$, initialized as ${\tau_c = \log(c/C)}$ for ${c=1,\ldots,C}$.
\item[--] A trainable $\tau$, initially drawn from the standard normal distribution $\mathcal N(0,1)$, trained with the gradient descent method along with the rest of the model parameters. 
\item[--] Using a separate network branch to compute values of $\tau$. We propose to use a small additional network branch inspired by the \emph{Squeeze-and-Excitation-nets} by Hu et al. \cite{hu}:
For each channel $x^c$ we compute the average 

\begin{equation}
\mu^c = \operatorname{GAP}(x^c) = \frac{1}{HW} \sum_{ij} x_{ij}^c.
\end{equation}

The vector $\mu$ of per-channel averages is then used to compute $\tau$ with

\begin{equation}
\tau = (\tau_1, \ldots, \tau_C) = F_2(\operatorname{ReLU}(F_1(\mu))).
\end{equation}
Here $F_1\colon \mathbb R^c \to \mathbb R^{C/r}$ and $F_2\colon \mathbb R^{C/r} \to \mathbb R^C$ are trainable affine maps, and $r$ is a reduction ratio. Each entry $\tau_i$ of $\tau$ is used for one channel.
We choose $r=16$ as proposed by Hu et al. \cite{hu}. (We assume that $r$ divides the number of channels evenly.)
\end{itemize}

\section{Experiments}

\subsection{Dataset}
\begin{figure}[htbp]
\begin{center}
\centerline{
\includegraphics[trim=0.4cm 0.2cm 0.4cm 0.4cm,clip,width=0.5\columnwidth
]{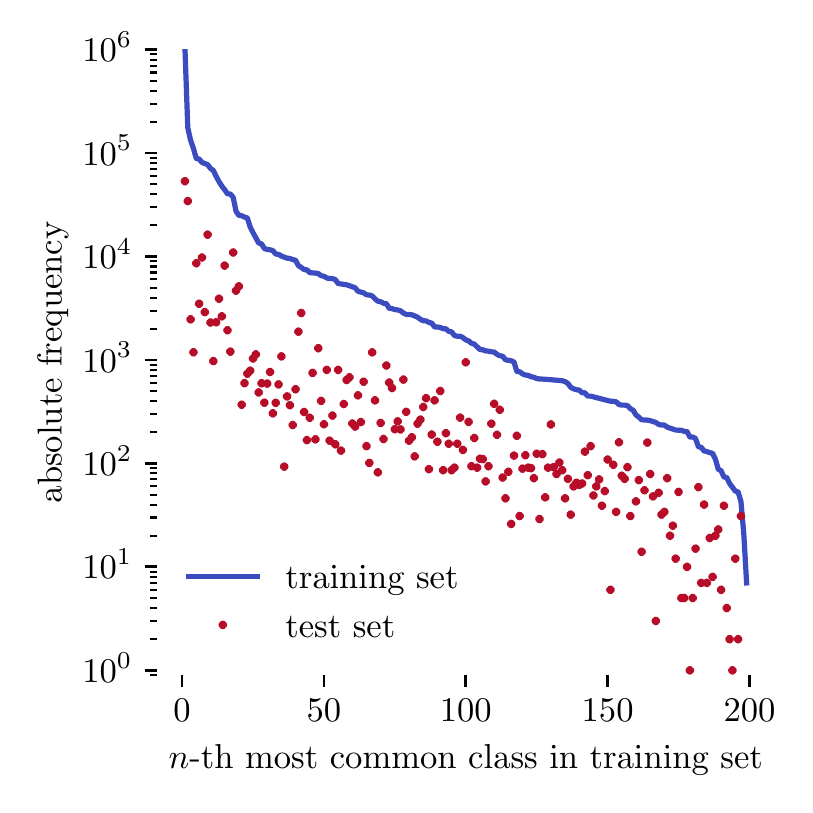}
}
\caption{Absolute frequency of each class, sorted by its frequency in the training set.}
\label{fig:frequencies}
\end{center}
\end{figure}

For all of our experiments, we train the networks for a classification task with 200 classes. The images along with their image-level labels are from the Google Open Images v5 dataset \cite{openImages}. (We used a subset of the images that were provided with bounding boxes, but only used the image-level labels.) This results in a training set of 2949053 images
and a test set of 220416 images. The frequency of each class is displayed in Figure \ref{fig:frequencies}.

\subsection{Image Augmentation}

For the training, the images are randomly rescaled such that the shorter edge has a length between $256$ and $384$. Then they are randomly cropped to a size of ${224 \times 224}$. For evaluating, the images are resized such that the shorter edges are always $256$ and then they are cropped at the center to again $224 \times 224$.
The range of the pixel values is mapped from $[0, 255]$ to $[0,1]$.

\subsection{Architecture}

To compare our proposed method to other downsampling- and pooling 
methods, we use a version of VGG16 \cite{zisserman}
with the setup that was described by Hu et al. \cite{hu}.
This includes \texttt{MP} as a baseline. We replace
the \texttt{MP} by the different aforementioned methods.

\begin{figure}
\centering
\begin{tabular}{cl}
Resolution & Blocks  \\
\hline 224 $\times$ 224 & C64, C64, Pool, \\ 
112 $\times$ 112 & C128, C128, Pool, \\
56 $\times$ 56 & C256, C256, C256, Pool, \\
28 $\times$ 28 & C512, C512, C512, Pool, \\
14 $\times$ 14 & C512, C512, C512, Pool, \\
7 $\times$ 7 & (512$\times$7$\times$7 to 25088), \\
 & FC4096, ReLU, Dropout(p=0.5), \\
 & FC4096, ReLU, Dropout(p=0.5), \\
 & FC200  \\
\end{tabular}
\captionof{table}{Here \texttt{CN} represents a block of $3\times 3$ convolutions of $N$ 
output channels followed by a batch normalization block and $\operatorname{ReLU}$ activation. 
\texttt{FCN} stands for an affine map with a codomain of dimension $N$. 
\texttt{Pool} is the place holder for the corresponding pooling block used in each of the experiments. The default is a \texttt{MP} block with $2\times2$ windows and a stride of $2$.}
\end{figure}

\subsubsection{Initialization}
We implemented the initialization of the convolutional- and fully-connected blocks just as in \cite{hu}. The convolutional weights are initialized with normally distributed weights according to He et al. \cite{kaiming}. In the affine blocks, the linear weights are populated with $\mathcal N(0,0.01)$ weights.

In the branch of our proposed smooth-maximum-pooling that computes the temperature parameters, we used uniformly distributed weights \cite{kaiming} initialization, just as in \cite{hu}. 

For each repetition of an experiment, we used a different seed to initialize the random number generators. The same set of seeds were used for every method.

\subsubsection{Pooling}
For consistency, we apply all pooling methods with ${2\times 2}$ windows and a stride of $2$, just like the reference. We compare the following methods: 

\begin{itemize}
\itemsep0em 
\item[--] Max-Pooling (\texttt{MP})
\item[--] Average-Pooling (\texttt{AP})
\item[--] Nearest-Neighbor-Downsampling (\texttt{NN})\\(Strided-Convolution) 
\item[--] Gated-Pooling (\texttt{GP})
\item[--] Ordinal-Pooling (\texttt{OP})
\item[--] Learned-Norm-Pooling (\texttt{LNP})
\item[--] Smooth-Max-Pooling with trainable temperature parameter (\texttt{SMP})
\item[--] Smooth-Max-Pooling with fixed temperature parameter (\texttt{SMPF})
\item[--] Smooth-Max-Pooling with Squeeze-and-Excitation-Block (\texttt{SESMP})
\item[--] Squeeze-and-Excitation-Block followed by Max-Pooling (\texttt{SEMP})\\ (We include this original configuration of the \emph{Squeeze-and-Excitation-nets} \cite{hu} as a reference.)
\end{itemize}

\subsubsection{Optimization}
For the training of the network we use the Adam optimizer by Kingma and Ba \cite{kingma} with the default values of ${(\beta_1, \beta_2) = (0.9, 0.999)}$ with a fixed learning rate of ${\operatorname{lr} = 10^{-4}}$, and a fixed number of $10$ epochs to eliminate any influence of additional hyperparameters.
We chose the learning rate by minimizing the classification error of the baseline over multiple runs with ${\operatorname{lr} = 10^{-1}, 10^{-2},\ldots,10^{-7}}$ of one epoch. The best value turned out to be ${\operatorname{lr} = 10^{-4}}$.  
For all other methods we confirmed this by testing the reduced set of values ${\operatorname{lr} = 10^{-3}, 10^{-4},10^{-5}}$.

\section{Results \& Discussion}

\subsection{Classification Performance} 
\begin{figure}[htbp]
\begin{center}
\centerline{
\includegraphics[trim=0in 0cm 0cm 0.0in,clip,width=0.5\columnwidth
]{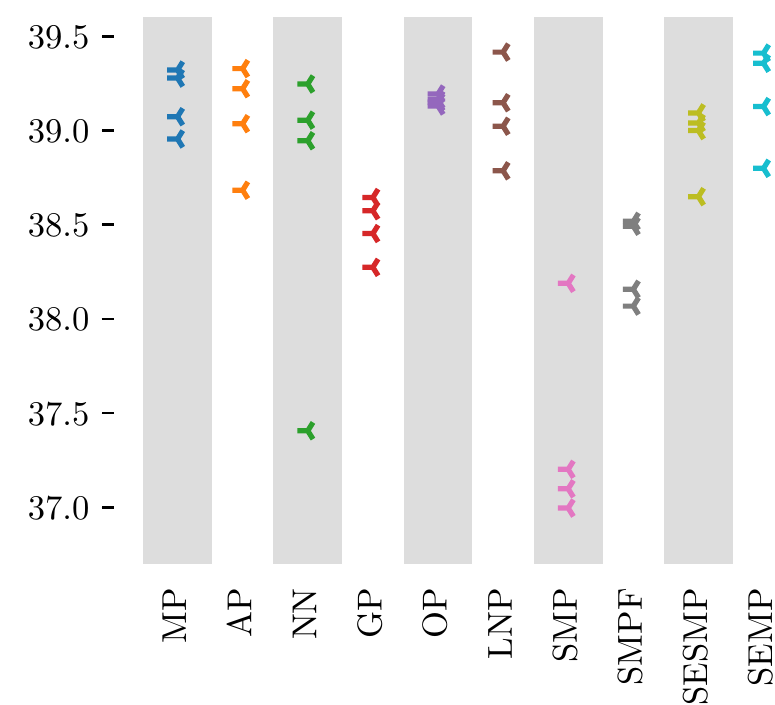}
}
\caption{Accuracy on the test set for each run (as a percentage).}
\label{fig:accuracies}
\end{center}
\end{figure}

We ran four training sessions for each pooling method that we tested to judge the consistency of the performance. 
Table \ref{tbl:accuracy} shows the average accuracy for each of the methods when evaluated on the test set for each of the runs, and in Figure \ref{fig:accuracies}, the accuracy on the test set is displayed for each run and method.

\begin{table}[t]
\caption{Accuracy on the training- and test- set as a percentage, averaged over the runs.}
\label{tbl:accuracy}
\begin{center}
\begin{small}
\begin{tabular}{lcc}
Method & Training & Test \\
\hline MP       &  65.69 $\pm$ \hphantom{0}.10 &  39.15 $\pm$ .14 \\
AP       &  65.69 $\pm$ \hphantom{0}.14 &  39.06 $\pm$ .24 \\
NN       &  63.94 $\pm$ 1.63 &  38.66 $\pm$ .73 \\
GP       &  63.34 $\pm$ \hphantom{0}.54 &  38.48 $\pm$ .14 \\
OP       &  65.75 $\pm$ \hphantom{0}.04 &  39.16 $\pm$ .02 \\
LNP      &  66.01 $\pm$ \hphantom{0}.19 &  39.09 $\pm$ .22 \\
SMP      &  60.82 $\pm$ 1.42 &  37.37 $\pm$ .47 \\
SMPF     &  64.16 $\pm$ \hphantom{0}.18 &  38.30 $\pm$ .19 \\
SESMP    &  65.00 $\pm$ \hphantom{0}.30 &  38.94 $\pm$ .17 \\
SEMP     &  65.83 $\pm$ \hphantom{0}.10 &  39.17 $\pm$ .24 \\
\end{tabular}
\end{small}
\end{center}
\end{table}

The first observation is that the classification accuracies of all runs and all pooling types are within less than $2.5$ percentage points on the test set. 
Furthermore, the three most popular methods (\texttt{MP, AP, NN}) 
all have a very similar performance, except for one outlier for \texttt{NN}.
\\
Most interesting is that the runs of \texttt{OP} exhibit a very small spread compared to the other methods.

For the \texttt{GP} method, we could see a slight decrease in performance over \texttt{MP}, 
even though Lee et al. \cite{lee} observed an increase. 
In their paper, the comparison was based on training on the MNIST \cite{mnist} 
and the CIFAR \cite{cifar} datasets, 
which contain only very small resolution images (${27 \times 27}$ and ${32 \times 32}$ respectively, where as we used a resolution of ${224 \times 224}$).

Similarly, we also could not find an improvement of 
performance using the methods we proposed. 
Both \texttt{SMP} and \texttt{SMPF} performed notably 
worse than other methods. Curiously, even though 
\texttt{SMPF} has no trainable parameters, 
it outperformed \texttt{SMP}. Out of these proposed 
methods \texttt{SESMP} performed best, and exhibited 
a similar accuracy as most of the other methods we tested.

\subsection{What did the pooling blocks learn?}
In \texttt{LNP}, \texttt{OP} and \texttt{SMP} the 
trained parameters indicate whether the behaviour 
is closer to \texttt{AP} or closer to \texttt{MP}. 
(For the other parametrized methods this is not 
possible: In these methods, the value of the interpolating 
parameter depends on the inputs.)

\paragraph{Learned-Norm-Pooling}

\begin{figure}[htbp]
\begin{center}
\centerline{
\includegraphics[trim=0.1in 0cm 0cm 0.25in,clip,width=0.5\columnwidth
]{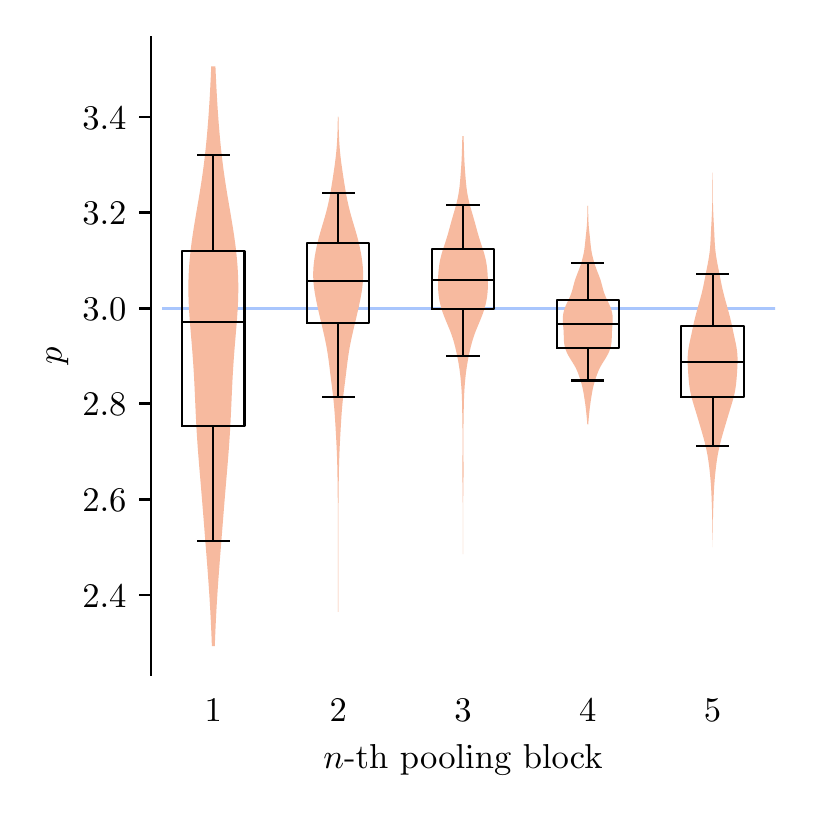}
}
\caption{Distribution of the value of $p$ of 
\texttt{LNP} across the channels  at the end 
of the training. The horizontal line represents the 
initial value. The box represents the median 
and the quartiles and the whiskers display 
the $5$-th and $95$-th percentile.}
\label{fig:lnp_params}
\end{center}
\end{figure}

The \texttt{LNP} method exhibited a similar performance 
in terms of accuracy as \texttt{AP} and \texttt{MP}. 
In Figure \ref{fig:lnp_params} we see that the value 
of the exponent $p$ of Equation \eqref{eq:flnp} 
changed significantly from the initial value. 
(Recall that the trainable parameter is $\tilde p$, 
and the value of the exponent $p$ is determined 
by Equation \eqref{eq:flnptilde}.) 
In the second and third pooling blocks, the exponents became 
more extreme, quite in contrast to the last block. 
This means the second and third blocks' behaviour 
shifted closer to \texttt{MP} while the last blocks' 
behaviour shifted closer to \texttt{AP}. 

However,
due to the way this method is parametrized, it is not
straightforward to make an absolute statement whether the behaviour is
closer to \texttt{AP} or \texttt{MP}. If we consider the
duality of $\ell^p$-spaces, one could claim that $p=2$ is 
half way between \texttt{AP} ($p=1$) and \texttt{MP} ($p=\infty$).
In that sense, all blocks could be considered closer to
\texttt{MP} than to \texttt{AP}. This observation, however, is  
skewed as the authors initialized with $p=3$~\cite{gulcehre}.

\begin{figure}[htbp]
\begin{center}
\centerline{
\includegraphics[trim=0.1in 0cm 0cm 0.2in,clip,width=0.5\columnwidth
]{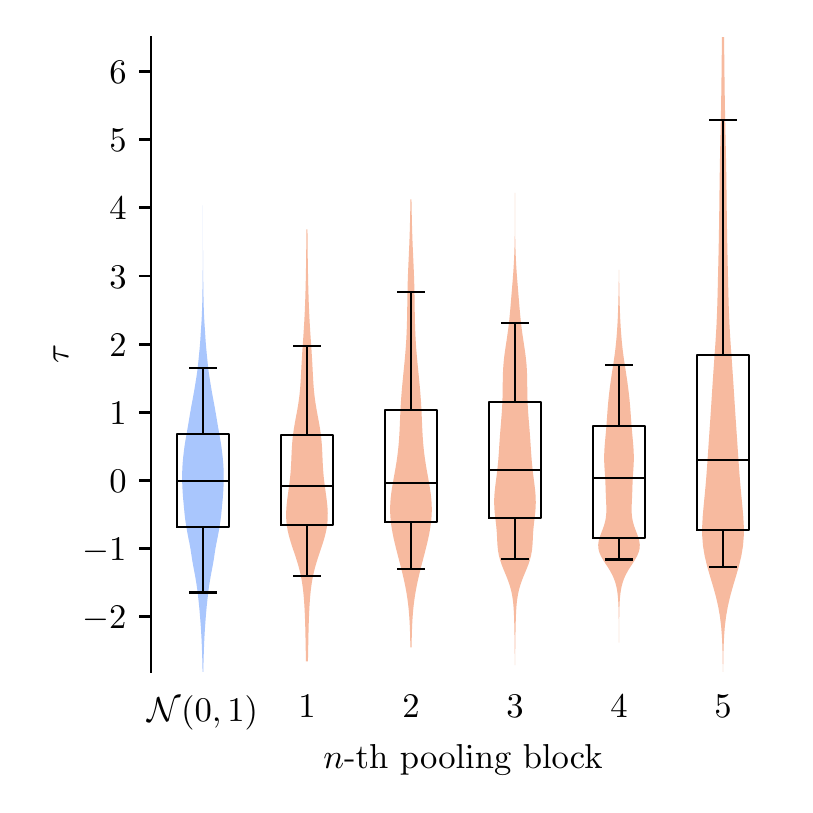}
}
\caption{Distribution of the value of $\tau$ of \texttt{SMP} 
across the channels at the end of the training. 
On the very left we show the box-plot of a standard 
normal distribution $\mathcal N(0,1)$ as a reference as the 
values of $\tau$ are initialized as realizations 
of $\mathcal N(0,1)$ at the start of the training. The box 
represents the median and the quartiles and the 
whiskers display the $5$-th and $95$-th percentile.}
\label{fig:smp_params}
\end{center}
\end{figure}

\paragraph{Smooth-Max-Pooling}
In \texttt{SMP} the temperature parameter $\tau$ (recall Equation \eqref{eq:fsmp}) 
was initialized by sampling from a standard normal distribution. 
In Figure \ref{fig:smp_params} we can observe a 
shift of $\tau$ in the positive direction, which 
means the general behaviour moved closer to 
\texttt{MP}. Still a large portion is centered 
around zero ($\tau = 0$ results in \texttt{AP}) 
and there are still some negative values present. We see 
that the distributions of parameters became assymetric with
relatively short in the direction of negative values and long 
tails in the direction of positive values.

Most prominent is the last pooling block, where 
the majority of values is positive, and the average 
magnitude of the positive values exceeds the one 
of the negative values by far. This tells us that 
there was a strong shift towards \texttt{MP}.

\begin{figure}[htbp]
\begin{center}
\centerline{
\includegraphics[trim=0.1in 0cm 0cm 0.15in,clip,width=0.5\columnwidth
]{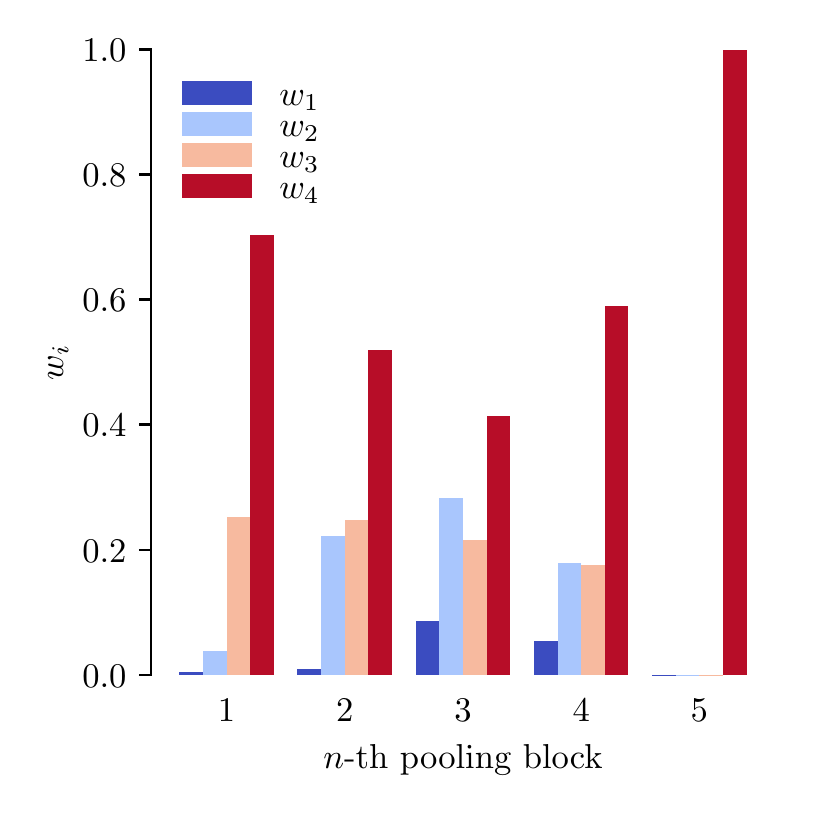}
}
\caption{The values of $w_i$ in \texttt{OP}. The value $w_1$ is the weight of the minimum, $w_4$ is the weight of the maximum.}
\label{fig:op_params}
\end{center}
\end{figure}

\paragraph{Ordinal-Pooling}
Recall that in \texttt{OP} the output is computed as a
 weighted average of the \emph{sorted} input values. 
 So the coefficient of the smallest number is $w_1$ and 
 the one of the largest is $w_4$. In Figure \ref{fig:op_params} 
 the values of these weights $w_i$ 
is displayed for each pooling block. 
In each block, the weight $w_4$ of the maximum exceeds 
the other weights, while $w_1$ is consistently below $0.1$. This indicates a strong shift towards \texttt{MP}. 
Just as in \texttt{SMP}, the most prominent is the 
very last block trained weights assumed almost the 
exact behaviour of \texttt{MP}. Out of all five blocks, the third one is closest to an \texttt{AP} block.

\bigskip
Both \texttt{SMP} and \texttt{OP} are symmetrical, as 
in theory, they could just as well converge to min-pooling 
as they can converge to max-pooling. In these experiments 
we saw that max-pooling was more strongly favoured than
min-pooling, which we explain with the presence of 
$\operatorname{ReLU}$ activation functions. These activation
functions clip negative values to zero, so one could argue
that min-pooling would result in irrelevant information
being passed forward.

\section{Conclusion}
We have shown that the choice of pooling functions does
not make a large difference when it comes to a classification problem.
Furthermore, all of the three most popular pooling methods - namely strided convolutions, 
\texttt{AP} and \texttt{MP} - perform similarly well and cannot easily be outperformed by more sophisticated downsampling methods.
Besides that, we showed that if the pooling blocks have 
the freedom to choose a behaviour, 
their favour more likely goes towards \texttt{MP}.

Lastly, we could only see a very small improvement over \texttt{MP} when applying Squeeze-and-Excitation networks (\texttt{SEMP}). 
It is insignificant when taking the spread of the different runs into account. Our experiments' limitation is that we used a fixed number of epochs for the training and a fixed learning rate. 
Using a more sophisticated learning rate schedule could benefit the training, but would make it more difficult to make a fair comparison.

What remains open is whether these results we presented 
for the VGG16 network also generalize to other classification 
architectures like the \emph{ResNet} \cite{resnet} or in the context of image segmentation networks like the \emph{U-Net} \cite{unet}, or even in discriminator networks of generative adversarial networks~\cite{gans}. 

Finally, there are still more tweaks that could be 
experimented with. One of these is, for example, 
the architecture of the branch that computes the 
temperature parameters in our proposed method.

\section{Appendix}

\subsection*{Numerically Stable Implementation of Smooth Maximum}
Note that scalar addition distributes over the smooth maximum:

\begin{equation}
f_{\operatorname{SMP}}^\tau(x_1,\ldots,x_n) + c  = f_{\operatorname{SMP}}^\tau(x_1+c,\ldots, x_n+c).
\end{equation}

Also, note that

\begin{equation}
f_{\operatorname{SMP}}^\tau(x_1,\ldots,x_n) = f_{\operatorname{SMP}}^1(\tau x_1,\ldots,\tau x_n).
\end{equation}

By choosing $d := \max_i \tau x_i$ we get 

\begin{equation}
f_{\operatorname{SMP}}^\tau(x_1,\ldots,x_n) 
 = f_{\operatorname{SMP}}^1(\tau x_1 - d,\ldots,\tau x_n-d)+d. \nonumber
\end{equation}

When evaluating the right-hand side, all arguments of the exponential function are in the range $(-\infty,0)$. This prevents the exponential function from overflowing. This is derived from the well known way to implement $\operatorname{softmax}(x)$ as $\operatorname{softmax}(x_1 - \max_i x_i, \ldots, x_n- \max_i x_i)$.

\paragraph{Gradients}

The smooth maximum is smooth, as the name suggests. And it is smooth in $x_i$ as well as in $\tau$. The derivative with respect to $\tau$ is

\begin{equation}
\frac{d f_{\operatorname{SMP}}^\tau(x)}{d \tau}
= \frac{
\sum_i x_i^2 e^{\tau x_i}}{\sum_i e^{\tau x_i}}
-
\Big(f_{\operatorname{SMP}}^\tau (x)\Big)^2 .
\end{equation}

This demonstrates another connection to the softmax: We already showed that if ${X \sim \operatorname{softmax}_\tau(x)}$ then ${f_{\operatorname{SMP}}^\tau(x) = E[X]}$, and furthermore, the equation above shows that ${f_{\operatorname{SMP}}^\tau(X) = E[X^2] - (E[X])^2 = V[X]}$, the variance of $X$. 

The derivative with respect to $x_i$ is

\begin{equation}
\frac{\partial f_{\operatorname{SMP}}^\tau(x)}{\partial x_i}
= \frac{(\tau x_i+1) e^{\tau x_i}  - \tau e^{\tau x_i}f_{\operatorname{SMP}}^\tau (x) }{\sum_j e^{\tau x_j}}.
\end{equation}

\end{document}